\documentclass[conference, 10 pt]{ieeeconf}  

\IEEEoverridecommandlockouts                              

\usepackage{subcaption}
\usepackage{cite}

\usepackage{graphicx} 
\usepackage{tikz}
\usepackage{pgfplots}
\usepackage{amsmath} 
\usepackage{amssymb}  
\usepackage{import}
\usepackage{dsfont}
\usepackage{hyperref}
\usepackage{gensymb}
\usepackage{booktabs}
\usepackage{tabularx}
\usepackage{diagbox}

\title{\LARGE \bf
A Multi-Task Recurrent Neural Network for End-to-End \\ Dynamic Occupancy Grid Mapping
}

\author{Marcel Schreiber$^{1}$, Vasileios Belagiannis$^{2*}$, Claudius Gl\"aser$^{3}$ and Klaus Dietmayer$^{1}$
	\thanks{The authors are with:}%
	\thanks{$^{1}$Institute of Measurement, Control, and Microtechnology, Ulm University, Germany, {\tt\footnotesize \{first.last\}@uni-ulm.de}}
	\thanks{$^{2}$Department of Simulation and Graphics, Otto von Guericke University Magdeburg, Germany, {\tt\footnotesize \{first.last\}@ovgu.de}}%
	\thanks{$^{3}$Robert Bosch GmbH, Corporate Research, 71272 Renningen, Germany, {\tt\footnotesize \{first.last\}@de.bosch.com}}%
	\thanks{*Most of this work was done while Vasileios Belagiannis was with Ulm University}
}
\begin{document}

\bstctlcite{IEEEexample:BSTcontrol} 

\def\cred{\textcolor{red}}
\def\cblue{\textcolor{blue}}
\def\cgreen{\textcolor{green}}

\newcommand\copyrighttextinitial{%

	\scriptsize This work has been submitted to the IEEE for possible publication. Copyright may be transferred without notice, after which this version may no longer be accessible.}%
\newcommand\copyrighttextfinal{%
	
	\scriptsize\copyright\ 2022 IEEE. Personal use of this material is permitted. Permission from IEEE must be obtained for all other uses, in any current or future media, including reprinting/republishing this material for advertising or promotional purposes, creating new collective works, for resale or redistribution to servers or lists, or reuse of any copyrighted component of this work in other works.}%
\newcommand\copyrightnotice{%

	\begin{tikzpicture}[remember picture,overlay]%

	\node[anchor=south,yshift=10pt] at (current page.south) {{\parbox{\dimexpr\textwidth-\fboxsep-\fboxrule\relax}{\copyrighttextfinal}}};%
	\end{tikzpicture}%


}

\newcommand{\ts}[1]{{\textsubscript{#1}}}
\newcommand{\tbf}[1]{{\textbf{#1}}}

\maketitle
\copyrightnotice%
\thispagestyle{empty}
\pagestyle{empty}

\begin{abstract}
A common approach for modeling the environment of an autonomous vehicle are dynamic occupancy grid maps, in which the surrounding is divided into cells, each containing the occupancy and velocity state of its location.
Despite the advantage of modeling arbitrary shaped objects, the used algorithms rely on hand-designed inverse sensor models and semantic information is missing.
Therefore, we introduce a multi-task recurrent neural network to predict grid maps providing occupancies, velocity estimates, semantic information and the driveable area.
During training, our network architecture, which is a combination of convolutional and recurrent layers, processes sequences of raw lidar data, that is represented as bird's eye view images with several height channels.
The multi-task network is trained in an end-to-end fashion to predict occupancy grid maps without the usual preprocessing steps consisting of removing ground points and applying an inverse sensor model.
In our evaluations, we show that our learned inverse sensor model is able to overcome some limitations of a geometric inverse sensor model in terms of representing object shapes and modeling freespace.
Moreover, we report a better runtime performance and more accurate semantic predictions for our end-to-end approach, compared to our network relying on measurement grid maps as input data.
\end{abstract}
%
%
\section{Introduction}
\begin{figure}
	\centering
	\includegraphics[width=0.98\columnwidth, trim=50 10 -20 -20, clip]{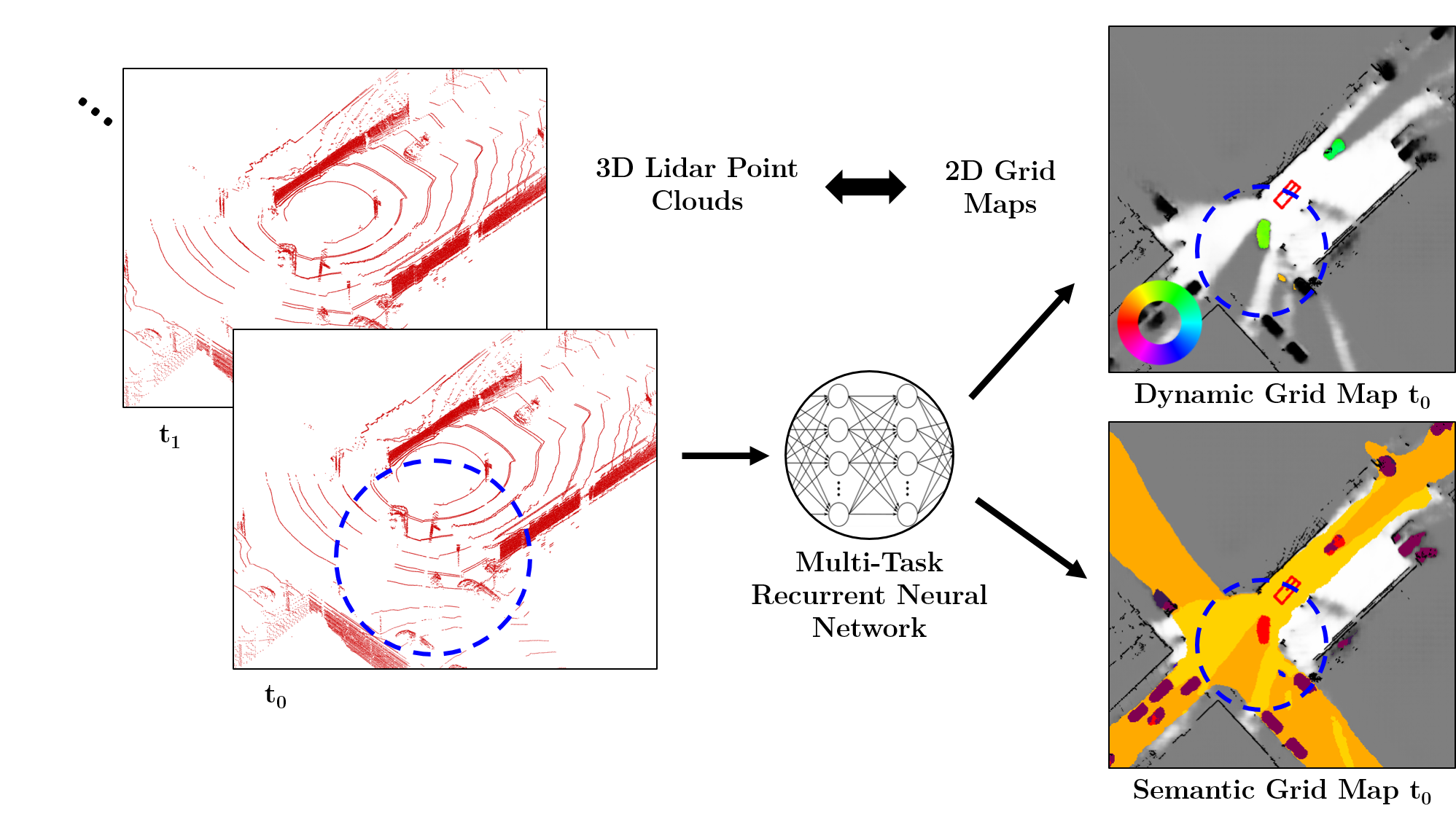}
	\caption{Illustration of our approach. The multi-task recurrent neural network uses the current 3D lidar point cloud and the recurrent states of the last time step to predict 2D grid maps.
	The visualizations show the occupancy probability in grayscale with occupied as black and free as white. 
	In the dynamic occupancy grid map, the colors are used to visualize the orientation of the velocity estimates. In the semantic grid map, the colors refer to semantic classes and the driveable area is visualized in yellow. The blue circle marks corresponding space in the 3D and 2D representation.
	}
	\label{fig:Teaser}
	\vspace{-0.6cm}
\end{figure}
For the realization of autonomous driving functions, a robust perception and modeling of the environment is a crucial prerequisite. 
A well-known representation for the surrounding of autonomous vehicles are dynamic occupancy grid maps, in which the environment is divided into equally sized cells, each containing the occupancy state of its location and a velocity estimate.
Several works \cite{TanzmeisterGridBasedMapping,Negre_HybridSamplingBayesian,DBLP:journals/corr/NussRTYKMGD16} use a particle-based algorithm, in which the cell state is represented by the set of particles located in it.
These algorithms are carried out by a prediction step, in which particles are able to move between cells, and an update step using an occupancy grid map, which is calculated by applying an inverse sensor model \cite{Thrun:2005:PR:1121596} using lidar data.  
The advantages of these dynamic occupancy grid maps are the ability to represent arbitrary shaped objects and estimate velocities, without the need for explicit object detection.
Thus, occupied cells can be divided into moving and static obstacles.
However, these grid maps do not provide further semantic information, for example whether an occupied cell belongs to a pedestrian or a vehicle.
Moreover, these approaches hardly rely on the inverse sensor model used for the calculation of the input data.
For this calculation, it is common to use a ground model to first remove the ground measurements, then the remaining lidar points are projected in a 2D grid. 
Cells with a lidar measurement are considered as occupied, cells between measurements and sensor are free, the remaining area is unknown.
%
This simple method has the drawback, that objects near the sensor could cause large occlusions, even though measurements are occurring in the space behind it. \\
In this work, we introduce a multi-task recurrent neural network to predict dynamic occupancy grid maps with semantic classes for the occupied cells, based on raw lidar data to tackle these drawbacks.
The multi-task network architecture is based on the approach in \cite{SchreiberDynamicOccupancyGridMapping}, but extended with an additional decoder for the semantic classification and omitting the calculation of measurement grid maps as input data. 
The input data in this work is a discretization of the 3D lidar data in a bird's eye view (BEV) with several channels to represent the height.  
The neural network performs the task to map these sparse 3D grids to a 2D occupancy grid map representation as visualized in Fig.~\ref{fig:Teaser} and therefore learns an inverse sensor model.
In addition to the prediction of an occupancy grid map, the multi-task network predicts the velocity and a semantic class for each occupied cell, and the driveable area.
In the evaluation, we show qualitatively that our learned inverse sensor model leads to an improved freespace estimation, compared to a classical inverse sensor model.
Moreover, we compare our end-to-end approach with prior work \cite{SchreiberDynamicOccupancyGridMapping} quantitatively and show benefits in the semantic classification.  
We also demonstrate a better runtime performance for the end-to-end approach, as fewer preprocessing steps are required. \\
To sum up, our contributions are \textit{1) the introduction of an end-to-end approach for occupancy grid mapping, omitting the need of a hand-designed inverse sensor model, 2) the prediction of an enhanced dynamic grid map consisting of occupancies, velocities, semantic classes and driveable area, 3) superior performance compared to prior work, that relies on measurement grid maps as input data.}
\section{Related Work}
\noindent \textbf{Dynamic occupancy grid maps} divide the surrounding of an agent in equally sized cells, each containing the occupancy state and a velocity estimate. 
In contrast to early approaches for occupancy grid maps \cite{ElfesOccGrids}, the dynamic occupancy grid maps are not restricted to the assumption of a static environment and are therefore well-suited in the context of autonomous driving.
Danescu et al.\cite{DanescuParticleBasedOccupancyGrid} propose to represent the occupancy and dynamic state of a cell with particles, that can move between cells.
Their approach is adopted in \cite{TanzmeisterGridBasedMapping} and \cite{Negre_HybridSamplingBayesian} by representing only the dynamic cells with particles, whereas Nuss et al.~\cite{DBLP:journals/corr/NussRTYKMGD16} formulate the dynamic occupancy grid mapping as a random finite set.
Despite these particle-based dynamic occupancy grid maps, there are several deep learning approaches for estimating velocities in a 2D bird's eye view (BEV) based on lidar data.
The work in \cite{SchreiberMotionEstimationInOccupancyGrids} propose to use a recurrent neural network architecture to estimate a dynamic occupancy grid map similar to \cite{DBLP:journals/corr/NussRTYKMGD16}, based on the same measurement grid maps as input data in a stationary setting.
In a subsequent work \cite{SchreiberDynamicOccupancyGridMapping} this approach is extended with an ego-motion compensation method for the application with a moving ego-vehicle.
The network architecture in~\cite{SchreiberDynamicOccupancyGridMapping} forms the basis of this work. 
It is extended to a multi-task network in order to additionally predict semantic classes and relies on different input data. 
Filatov et al.~\cite{AnyMotionDetector} introduce a recurrent network architecture for the prediction of a velocity grid based on a sequence of lidar point clouds. 
First, the lidar data is processed with a voxel feature encoding layer \cite{Zhou2018VoxelNetEL} to obtain bird's eye view representations, which are aggregated with a convolutional recurrent network layer. The output of this recurrent layer is then used in a ResNet-18-FPN backbone to predict the velocity grid. 
Lee et al. \cite{lee2020pillarflow} propose to use pillar feature networks \cite{lang2019pointpillars} to encode two consecutive lidar sweeps in a bird's eye view image. These feature maps are then processed in a feature pyramid and a flow network to predict the scene flow in a 2D grid.
Compared to dynamic occupancy grid maps, the work in \cite{AnyMotionDetector} and \cite{lee2020pillarflow} focus on the velocity estimation, but do not model freespace and occlusions. 
Wu et al. \cite{MotionNet} introduce a spatio-temporal network architecture to predict a 2D grid encoding motion and semantic for each cell, based on 3D lidar point clouds.
The lidar data is encoded in a binary 3D grid, in which cells with at least one lidar measurement are determined as occupied and processed as a 2D bird's eye view image with the height dimension as image channels.
In this work, we use a similar input representation, but with a larger spatial size and a finer discretization.
The output representation in \cite{MotionNet} differs from this work, as they classify grid cells containing at least one lidar measurement, but do not model the freespace or occlusions. \\
\noindent\textbf{Inverse sensor models} are used in the context of occupancy grid maps to describe the occupancy state of a grid cell based on a given sensor measurement~\cite{ElfesOccGrids, Thrun:2005:PR:1121596}.
In geometric inverse sensor models (ISM) for range measurements, e.g. lidar sensors, grid cells with a lidar reflection are assigned as occupied, the space between sensor and reflection as free, the succeeding cells are considered as unknown. 
The accuracy of such a model depends on the preprocessing step to remove ground measurements.
Additionally, it has limitations, for example in situations with lidar measurements behind an object, as this is simply considered as occluded space.
%
An alternative is to use learned sensor models that take raw sensor data as input to a neural network that predicts occupancy probabilities as first proposed in \cite{ThrunExplorationAndModelBuilding} for building an occupancy map based on sonar range sensors.
Dequaire et al. \cite{DequaireDeepTrackingInTheWild} propose an end-to-end trainable recurrent neural network using simple occupancy grid maps as input data to predict occupied cells in unobserved area. Their model is also extended to predict semantic classes for occupied cells.
The approach in \cite{WirgesEvidentialOccupancyGridMapAugmentation} use multiple lidar scans to generate occupancy maps with an extended field of view and less occlusions, that serve as label data.
They use a multi-layer grid, encoding the detections, transmissions and intensities from a single lidar scan as input for a neural network to predict these augmented occupancy grid maps.
A deep learning based framework for occupancy grid mapping is introduced in \cite{VanKempenASimulationBasedEndToEndLearning}.
They generate training data from simulation, which provides an accurate ground truth occupancy grid map.
These maps are augmented with a ground truth object list by setting cells inside the bounding boxes to occupied.
An extended point pillar network architecture \cite{lang2019pointpillars} is used to predict these grid maps, based on a single lidar scan.
In this work, we aim to predict a dynamic occupancy grid map without applying an inverse sensor model.
This approach differs from \cite{VanKempenASimulationBasedEndToEndLearning} as we rely on recurrent neural networks in order to use information of several time steps for the prediction of occupancy and velocity. 
Additionally, we provide semantic classes and the prediction of the driveable area.\\
\noindent \textbf{Semantic grid maps} describe the environment in a 2D grid, where each cell contain semantic information of the space it is located. 
Several works~\cite{LuMonocularSemanticOccupancyGridMapping,RoddickPredictingSemanticMapRepresentations} propose to predict 2D semantic occupancy grid maps based on image data using an end-to-end trainable network architecture.
Erkent et al.\cite{ErkentSemanticGridEstimationWithAHybrid} combine an occupancy grid map based on a Bayesian filter method \cite{RummelhardCMCDOT} with an image-based semantic segmentation using a fusion network.
These works have the advantage of rich semantic information in rgb image data, but encounter the problem of accurately mapping the data from the image domain to a grid structure.
Bieder et al.~\cite{BiederExploitingMultiLayerGridMapsForSurroundViewSemantic} propose to transform a semantically annotated 3D point cloud into a 2D grid map representation and fuse static semantic information of several time steps to obtain dense semantic grid map labels.
These labels are then used to train a fully convolutional network, based on DeepLabV3 \cite{DeepLab} relying on multi-layer grid maps with handcrafted features as input data.
In a following work \cite{BiederImprovingLidarBasedSemanticSegmentationOfTopViewGridMaps}, they propose to process a spherical 2D projection with RangeNet++~\cite{MilitoRangeNet++} and fuse this semantic segmentation with a multi-layer grid map in a DeepLabV3 \cite{DeepLab} based network architecture.
Both approaches \cite{BiederExploitingMultiLayerGridMapsForSurroundViewSemantic,BiederImprovingLidarBasedSemanticSegmentationOfTopViewGridMaps} use a multi-layer grid with handcrafted features as input data and the same semantic grid representation as output. 
Relying on the same label data, Fei et al.~\cite{FeiPillarSegNet} propose to generate pillar features~\cite{lang2019pointpillars} and fuse them with an observability map as input for an encoder-decoder network.
Their approach receive improved results compared to \cite{BiederImprovingLidarBasedSemanticSegmentationOfTopViewGridMaps}, especially for small objects.
The difference of the dense semantic grid maps predicted in \cite{BiederExploitingMultiLayerGridMapsForSurroundViewSemantic, BiederImprovingLidarBasedSemanticSegmentationOfTopViewGridMaps, FeiPillarSegNet} compared to a dynamic occupancy grid map is the missing information whether obstacles are static or moving as well as the modeling of occlusions caused by dynamic objects.
In this work, we combine the prediction of a dynamic occupancy grid map with the semantic classification of occupied cells, based on lidar input data.
\section{Methods}\label{sec:methods}
In this section, we present our multi-task recurrent network architecture to predict dynamic occupancy grid maps with semantic classes in an end-to-end fashion. 
The proposed system relies on sequences of lidar point clouds, which are represented in a simple bird's eye view representation.
The label for network optimization are occupancy grid maps, that are augmented with the extent, velocity and semantic class of objects.
Additionally, the classification of cells with at least one lidar measurement as ground is used as auxiliary task to improve the network predictions.
An overview of the approach is shown in Fig.~\ref{fig:SystemOverview}. 
\subsection{Input Data - Point Cloud Projections}\label{sec:InputData}
In order to apply standard 2D convolutions on lidar point clouds, we represent the input data as BEV images, similar as in \cite{YangPIXOR} for 3D object detection, but without using the lidar intensity feature.
The input for our network is a 3D tensor with the size $L\times W \times H$, holding the binary occupancy state, i.e. a one if there is at least one lidar point located in a 3D grid cell, otherwise zero.   
For the calculation, we first define a grid map with $L \times W$ cells in the x-y-plane and the grid resolution $d\ts{l} \times d\ts{w}$ with the ego-vehicle located in the middle cell. 
The number of the channels in the z-dimension $H$ depends on the minimum height $h\ts{min}$, maximum height $h\ts{max}$ and the cell height $d\ts{h}$
\begin{equation}
	H = \left\lceil \frac{h\ts{max} - h\ts{min}}{d\ts{h}} \right\rceil + 2.
\end{equation}
Note, that two channels are added to set the occupancy state for out-of-range lidar points.
Assuming, that the lidar points $\{l_i=(x_i, y_i, z_i)\}^{N}_{i=1}$ are received in the vehicle coordinate system, the tensor indices $j\ts{x}, j\ts{y}, j\ts{z}$ can be received with
\begin{equation}
\begin{aligned}
j\ts{x} &= \left\lfloor \frac{x}{d\ts{l}} + \frac{L}{2} \right\rfloor \\
j\ts{y} &= \left\lfloor \frac{y}{d\ts{w}} + \frac{W}{2} \right\rfloor \\	
j'\ts{z}&= \left\lfloor \frac{z - d\ts{min}}{d\ts{h}} \right\rfloor + 1 \\
j\ts{z} &= \text{min}(\text{max}(j'\ts{z},0),H-1),\\ 
\end{aligned}
\label{eq:tensorIndex}
\end{equation}
ignoring points with $j\ts{x}$ or $j\ts{y}$ out of bounds.
Thus, for input calculation, we simply iterate over the lidar points and set the entry at the associated tensor index to one.
Before the association to the tensor index in (\ref{eq:tensorIndex}), we apply a rotation to a global coordinate system, in order to apply the ego-motion compensation method proposed in \cite{SchreiberDynamicOccupancyGridMapping}.
The described input representation seems suitable for predicting occupancy grid maps, as the lidar data is mapped to the same 2D grid dimension and resolution as at the network output. 
Additionally, no lidar points are removed, unless there are multiple measurements in one 3D grid cell, which is important as the method makes use of sparse ground points to estimate the freespace.
\subsection{Input Data - Measurement Grid Maps}\label{sec:MeasGrids}
\begin{figure}
	\centering
	\vspace{1.42mm}
	\includegraphics[width=0.49\columnwidth]{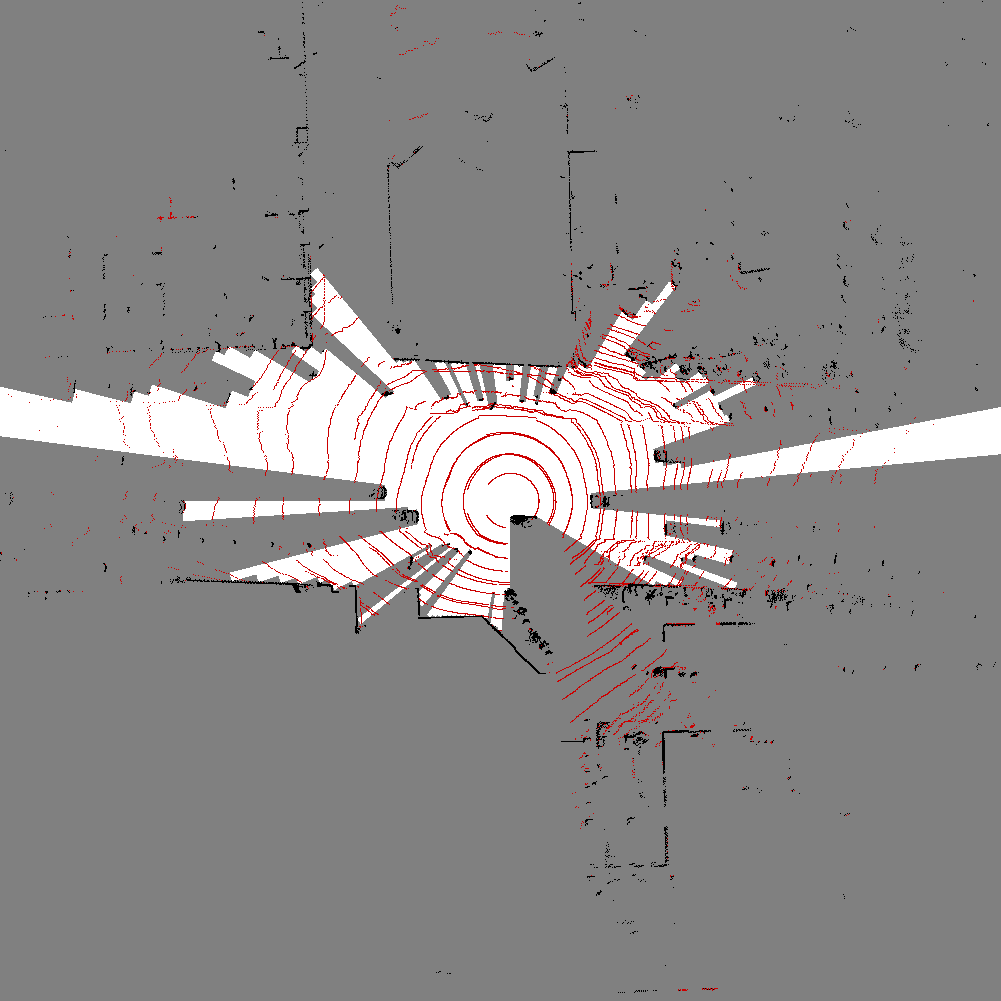}
	\includegraphics[width=0.49\columnwidth]{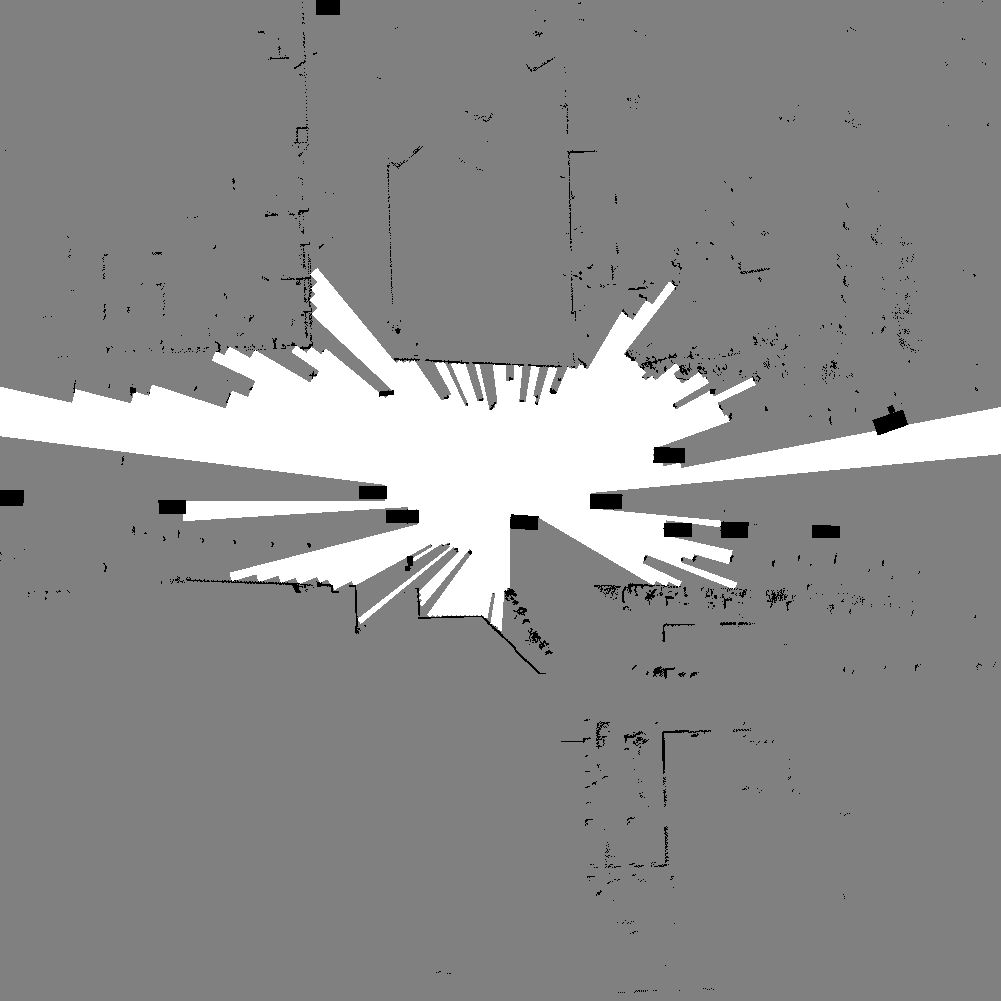}
	\caption{Illustration of the measurement grid map (left) and the occupancy label augmented with the bounding boxes of object labels (right). Occupied cells are visualized in black, freespace in white and unknown area in gray. In the left image the lidar ground points are visualized in red, highlighting the limitations of the geometric inverse sensor model.}
	\label{fig:MeasGrid}
	\vspace{-0.5cm}
\end{figure}
A grid map divides the vehicle environment in equally spaced cells, where each cell holds information for the space it is located.
In occupancy grid maps each cell contain the occupancy state $o_k$, which determine whether this space is occupied or free at the discrete time step $k$ \cite{ElfesOccGrids, Thrun:2005:PR:1121596}.
Measurement grid maps are occupancy grid maps that are based on lidar data of only one single time step. 
In this work, we calculate measurement grid maps using a simple geometric inverse sensor model.
First, the ground points in the lidar data are removed using map information, then the remaining lidar points are projected to the 2D grid map and each cell containing a lidar point is set to occupied. 
Cells between lidar reflections and sensor are set to free, cells in occluded space or outside the sensors field of view are determined as unknown.
These measurement grid maps serve as basis for our occupancy grid map label.
In addition, we use it as network input in our baseline experiment for comparison to our new approach relying on the input data described in Section \ref{sec:InputData}.
An example of these measurement grid maps is shown in Fig.~\ref{fig:MeasGrid} on the left, with the ground points visualized in red.
The depicted scene shows a drawback of this simple geometric inverse sensor model, as some space with ground measurements is falsely determined as unknown.
\subsection{Network Outputs and Labels}\label{sec:LabelData}
\begin{figure*}
	\centering
	\includegraphics[width=\textwidth]{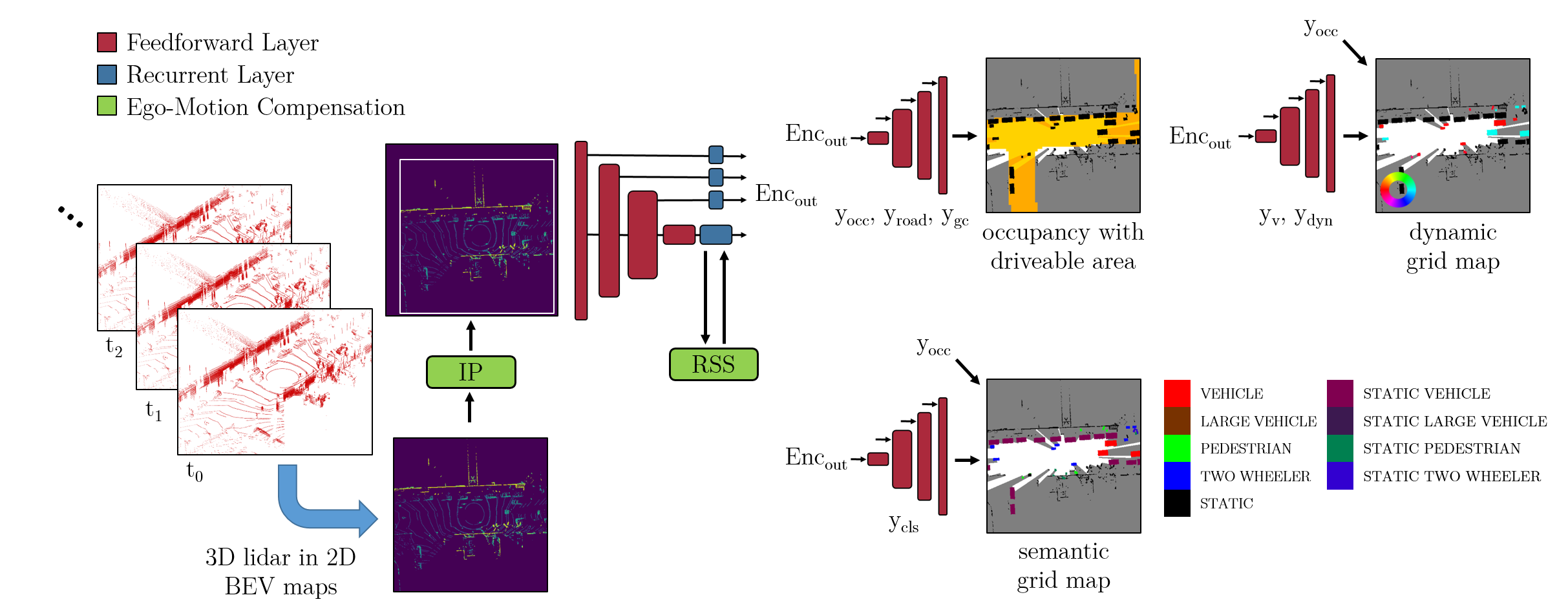}
	\caption{Illustration of our multi-task approach. The network architecture consists of a feedforward encoder and three feedforward decoders visualized in red. During training a sequence of lidar data represented as binary 2D BEV maps are processed with the encoder and aggregated in the states of the recurrent network layers (blue). The network jointly performs the prediction of the occupancy $y\ts{occ}$, driveable area $y\ts{road}$, velocities $y\ts{v}$ and semantic classes $y\ts{cls}$. The classification of ground points $y\ts{gc}$ and the dynamic cells $y\ts{dyn}$ are auxiliary tasks. The visualizations on the right show the label data. The occupancy label is visualized in grayscale, with the occupied area in black and freespace in white, additionally the driveable area label is depicted in yellow. 
	In the dynamic grid map, colors are used to visualize the orientation of the velocities for dynamic cells, in the semantic grid map for the semantic classes according the legend.}
	\label{fig:SystemOverview}
	\vspace{-0.5cm}
\end{figure*}
The base configuration of our multi-task network outputs an occupancy grid map $y\ts{occ}$, the velocities $y\ts{v}$, the prediction of dynamic cells $y\ts{dyn}$ and a semantic classification $y\ts{cls}$ for the occupied cells.
The network can be further extended with the prediction of the driveable area $y\ts{road}$ and an auxiliary task $y\ts{gc}$ to classify cells with a lidar point as ground cells.
Each network output is a 3D tensor, i.e. a grid map with one or more layers, with the same spatial size $L \times W$.
We use the \textit{Argoverse 3D Tracking Dataset}~\cite{Argoverse}, which provides lidar data, annotated 3D object boxes and map data to calculate the labels for the aforementioned tasks.
For the occupancy label, we use the measurement grid maps, described in Section~\ref{sec:MeasGrids}, but augment these grid maps using annotated object labels to set all cells inside a bounding box to occupied as shown in Fig.~\ref{fig:MeasGrid}.
The velocity and semantic labels for these occupied areas, can then be set according the object label.
The ground truth velocities are calculated from the displacement of the object boxes in a global coordinate frame and stored as a velocity pointing east $v\ts{E}$ and north $v\ts{N}$.
For the prediction of dynamic area, all cells with a velocity magnitude $v\ts{mag} > 0.8$~m/s are defined as dynamic.
The semantic labels are a coarser subset of the class labels from the \textit{Argoverse 3D Tracking Dataset}~\cite{Argoverse} as shown in Table~\ref{tab:ArgoverseClassMapping}.
\begin{table}
	\centering
	\vspace{1.42mm}
	\caption{Mapping to coarser classes}
	\label{tab:ArgoverseClassMapping}
	\begin{tabular}{@{}c|c@{}}
		\toprule
		grid map classes & argoverse classes \\
		\midrule
		\midrule
		VEHICLE (V) & VEHICLE \\
		\midrule
		LARGE VEHICLE (LV) & LARGE VEHICLE, SCHOOL BUS, \\
		& EMERGENCY VEHICLE, TRAILER \\
		\midrule
		PEDESTRIAN (PED) & PEDESTRIAN, STROLLER, ANIMAL, \\
		& WHEELCHAIR \\
		\midrule
		TWO WHEELER (TW) & BICYCLE, BICYCLIST, MOPED, \\
		& MOTORCYCLE, MOTORCYCLIST \\
		\midrule
		STATIC (STA) & ON ROAD OBSTACLE, OTHER MOVER, \\
		& occupied cells without label \\
		\bottomrule
	\end{tabular}
	\vspace{-0.5cm}
\end{table}
We choose to use coarser labels, as some classes are hardly to distinguish in lidar scans and especially in measurement grid maps, which serve as input in our baseline experiment.
Additionally, we split the class labels in dynamic and static classes.
We argue, that it is relatively easy for the recurrent network architecture to classify moving objects compared to distinction between a small static object and the static environment.
The \textit{Argoverse 3D Tracking Dataset}~\cite{Argoverse} provides binary driveable area labels as rasterized map with a resolution of one meter, which we use to generate labels in the spatial size and resolution of our grid map data.
The argoverse map data is also used to split the lidar point cloud in ground points and no ground points, which is then used to generate labels that associate each cell to one of the classes \textit{\{BACKGROUND, GROUND, NO GROUND\}}.
These labels are used for the auxiliary task $y\ts{gc}$ to improve the network optimization.
In Fig.~\ref{fig:SystemOverview} the labels for the multiple outputs are visualized.
\subsection{Multi-Task Recurrent Network Architecture}
The multi-task network visualized in Fig.~\ref{fig:SystemOverview} is based on the architecture in \cite{SchreiberDynamicOccupancyGridMapping}, but extended with additional tasks. 
Additionally, the architecture relies on lidar data instead of measurement grid maps as input.  
The network is a combination of a U-Net \cite{DBLP:journals/corr/RonnebergerFB15} based feedforward architecture and recurrent layers on each network level.
During training a sequence of input data in $\mathds{R}^{n\ts{in} \times W' \times L' \times H}$ is processed through several convolutional layers, reducing the spatial size three times by a factor of three to $\mathds{R}^{n\ts{in} \times \frac{W'}{27} \times \frac{W'}{27} \times 128}$.
This tensor is used as input for a convolutional long short-term memory (ConvLSTM)~\cite{DBLP:journals/corr/ShiCWYWW15}.
Additionally, there are \mbox{ConvLSTMs} in the skip connections to use temporal information on several scales.
The outputs of the recurrent layers serve as input for three decoders. The first decoder is used to predict the occupancy $y\ts{occ}$, the driveable area $y\ts{road}$ and perform the ground classification $y\ts{gc}$ as auxiliary task.
The second decoder predicts the velocities $y\ts{v}$ for occupied cells and perform the classification of dynamic cells $y\ts{dyn}$ as auxiliary task.
Finally, the third decoder performs a semantic classification $y\ts{cls}$ for occupied cells. 
The ego-motion compensation consists of the input placement (IP) and the recurrent states shifting (RSS) as introduced in \cite{SchreiberDynamicOccupancyGridMapping}.
This method allows to use recurrent layers on different network levels, e.g recurrent states with different resolutions.
The IP is used to compensate small movements of the ego-vehicle at the input of the network. 
This is achieved by increasing the input data $W' = W + 28, L' = L + 28$ with an adjustable padding operation inside the network graph. 
In order to compensate larger movements over several time steps, all recurrent states are shifted synchronously in the RSS module, if the ego-vehicle moves between cells in the most inner network level.
\subsection{Loss Functions}
The optimization of the multi-task network relies on the minimization of the summation of all task specific loss terms
\begin{equation}
\begin{aligned}
	L\ts{o}=& \alpha\ts{occ} L\ts{occ} + \alpha\ts{v} L\ts{v} + \alpha\ts{dyn} L\ts{dyn} \\ 
	&+ \alpha\ts{road} L\ts{road} + \alpha\ts{cls} L\ts{cls} + \alpha\ts{gc} L\ts{gc}
\end{aligned}
\end{equation}
with weight factors ${\alpha\ts{occ}=5}$, ${\alpha\ts{v}=0.02}$, ${\alpha\ts{dyn}=0.1}$, ${\alpha\ts{road}=1}$, ${\alpha\ts{cls}=2}$, ${\alpha\ts{gc}=1}$ for each task.
The loss functions for the occupancy $L\ts{occ}$, the velocities $L\ts{v}$, the prediction of the dynamic area  $L\ts{dyn}$ and driveable area  $L\ts{road}$ are defined as 
\begin{equation}
	L\ts{*} = \frac{1}{L\cdot W} \sum_{i=1}^{L\cdot W} \lambda\ts{*}(i)(y\ts{*}(i)-\hat{y}\ts{*}(i))^2
\end{equation}
with $* = \{\text{occ, v, dyn, road}\}$, the prediction $\hat{y}\ts{*}(i)$, the ground truth $y\ts{*}(i)$ and the task specific weighting $\lambda\ts{*}(i)$ of cell $i$.
The cell-wise weightings $\lambda\ts{dyn}$ and $\lambda\ts{v}$ are used to encounter the imbalance between dynamic and static area.
We set them both to 20 for dynamic cells, 5 for static cells and zero for all not occupied cells, as the predictions are only valid for occupied cells, i.e. cells with $p\ts{o}>0.7$.
For the optimization of the driveable area prediction, we use the measurement grid maps to distinguish between observable area $\lambda\ts{road} = 1$ and not observable area $\lambda\ts{road} = 0.25$.
We choose a lower weighting for the unobservable area, otherwise the network might start hallucinating too much driveable area.
The output layers for the semantic and ground classification consist of a softmax activation to normalize the class scores to class probabilities.
The loss function for the semantic classification $L\ts{cls}$ and the ground classification $L\ts{gc}$ is a cross-entropy loss
\begin{equation}
	L\ts{CE} = -\frac{1}{L\cdot W} \sum_{i=1}^{L\cdot W}\sum_{c=1}^{C} \lambda\ts{CE}(c,i)y(c,i)log(\hat{y}(c,i))
\end{equation}
with a cell-wise weighting per class $\lambda\ts{CE}$, the prediction $\hat{y}$ and the ground truth $y$ for $C=9$ classes for the semantic classification and $C=3$ classes for the ground classification.
For the ground classification, we set $\lambda\ts{CE} = 0.1$ for the background class, otherwise $\lambda\ts{CE} = 0.5$.
For the semantic classification, we use the focal loss as proposed in \cite{LinFocalLoss} to set $\lambda\ts{CE}$. 
As the semantic classification is only performed for the occupied cells, we set $\lambda\ts{CE} = 0$ for the not occupied cells.
\section{Experimental Setup}
In our experiments, we employ the \textit{Argoverse 3D Tracking Dataset}\cite{Argoverse} to generate training data as described in Section \ref{sec:LabelData} and to evaluate our approach. In the following, we introduce this dataset briefly and provide some implementation details.
\subsection{Dataset}
The \textit{Argoverse 3D Tracking Dataset}\cite{Argoverse} consists of 113 sequences, each with a length of 15 to 30 seconds.
In our work, we use the 65 training sequences for network optimization and the 24 validation sequences for evaluation, as there are no labels provided for the 24 test sequences.
Our input data relies on the lidar data, which is recorded with two roof-mounted lidar sensors with 32 beams each and a range up to 200~m.
The dataset provides annotated 3D bounding boxes with semantic class labels and labels for the driveable area, which are used to generate our ground truth data as described in Section \ref{sec:LabelData}.
In the label data generation, we choose a grid cell size of 0.15~m and a grid map size with $1001\times1001$ cells, leading to a perception area of [-75m, 75m].
The dataset provides object labels only 5 meters beyond the driveable area, which is the region of interest (ROI). 
Note, that the tasks $y\ts{occ}$, $y\ts{cls}$, $y\ts{v}$ and $y\ts{dyn}$ are only optimized using this ROI, as the labels for these tasks depend on the object labels.
\subsection{Implementation Details}
For the optimization of our network, we use the adam optimizer \cite{DBLP:journals/corr/KingmaB14} with an initial learning rate of 0.0001, which we divide by a factor of 2 every 100k training iterations. 
During training, we rely on input sequences with $n\ts{in}=10$ time steps and use the predictions of the last two time steps for the loss calculation.
In addition, we rotate the training samples randomly in $1\degree$ steps and apply dropout in the non-recurrent connections of the ConvLSTM layers as proposed in \cite{DBLP:journals/corr/ZarembaSV14}.
Due to memory constrains, we train our models using grid maps with $601\times601$ cells. 
Note, that for our evaluation, we use the full size grid maps with $1001\times1001$ cells.
For our input data, we choose the minimum height $h\ts{min}=-1.6~\text{m}$, the maximum height $h\ts{min}=3.0~\text{m}$ and the discretization $d\ts{h} = 0.2\text{m}$, which leads to 25 height channels.
\section{Evaluation}
\begin{table}[]
	\vspace{1.42mm}
	\renewcommand{\arraystretch}{1.2}
	\centering
	\caption{Different network configurations. The two input options are the point cloud projection (BEV) or the measurement grid map (MGM).}
	\label{tab:Configurations}
	\setlength{\tabcolsep}{0.1cm}
	\begin{tabular}{|c||cc||cccccc|}
		\hline
		 & \multicolumn{2}{c||}{input data}  & \multicolumn{6}{c|}{outputs}     \\ \hline
		method & BEV & MGM & $y\ts{occ}$ & $y\ts{v}$ & $y\ts{dyn}$ & $y\ts{cls}$ & $y\ts{road}$ & $y\ts{gc}$  \\ \hline
		meas\_base & & $\checkmark$ & $\checkmark$ & $\checkmark$ & $\checkmark$ & $\checkmark$ & & \\ 
		lidar\_base & $\checkmark$ &  & $\checkmark$ & $\checkmark$ & $\checkmark$ & $\checkmark$  &  & \\ 
		lidar\_base\_gc & $\checkmark$  &  & $\checkmark$ & $\checkmark$ & $\checkmark$ & $\checkmark$  &  & $\checkmark$  \\ 
		lidar\_base\_road & $\checkmark$  &  & $\checkmark$ & $\checkmark$ & $\checkmark$  & $\checkmark$  & $\checkmark$  & \\ 
		lidar\_base\_road\_gc & $\checkmark$  &  & $\checkmark$ & $\checkmark$ & $\checkmark$ & $\checkmark$  & $\checkmark$  & $\checkmark$ \\ \hline
	\end{tabular}
	\vspace{-0.5cm}
\end{table}
The evaluation relies on the 24 validaton sequences of the \textit{Argoverse 3D Tracking Dataset} \cite{Argoverse}.
The base configuration of our network \textit{lidar\_base} provides the prediction of occupancy, velocities with the prediction of dynamic area and the semantic classification as shown in Table~\ref{tab:Configurations}.
We train the same configuration using measurement grid maps as input data, denoted as \textit{meas\_base}.
Note, that \textit{meas\_base} has the advantage, that the occupancy label is equal with the input data, but augmented with object boxes. 
So, the approach \textit{meas\_base} use input data with already error-free removed ground points and only needs to learn the augmentation with the object shapes.
In contrast, the \textit{lidar\_base} approach has to detect ground points internally to predict the occupancy grid, i.e. learn an inverse sensor model.
First, the occupancy predictions of this two approaches are evaluated qualitatively.
Second, we evaluate the ability to separate occupied cells in static and dynamic, and the accuracy of the velocity predictions.
Furthermore, the influence of the input data on the semantic classification performance is investigated.
In the quantitative evaluation, we also evaluate the \textit{lidar\_base} approach with the driveable area prediction and the ground classification added.
Finally, we report runtime performances and show some qualitative results of our approach on data from our autonomous driving platform.
\subsection{Representation of Occupancy and Freespace}
\begin{figure}
	\vspace{1.42mm}
	\centering
	\begin{subfigure}{0.32\columnwidth}
	\includegraphics[trim=10cm 13cm 10cm 6cm, clip, width=\columnwidth]{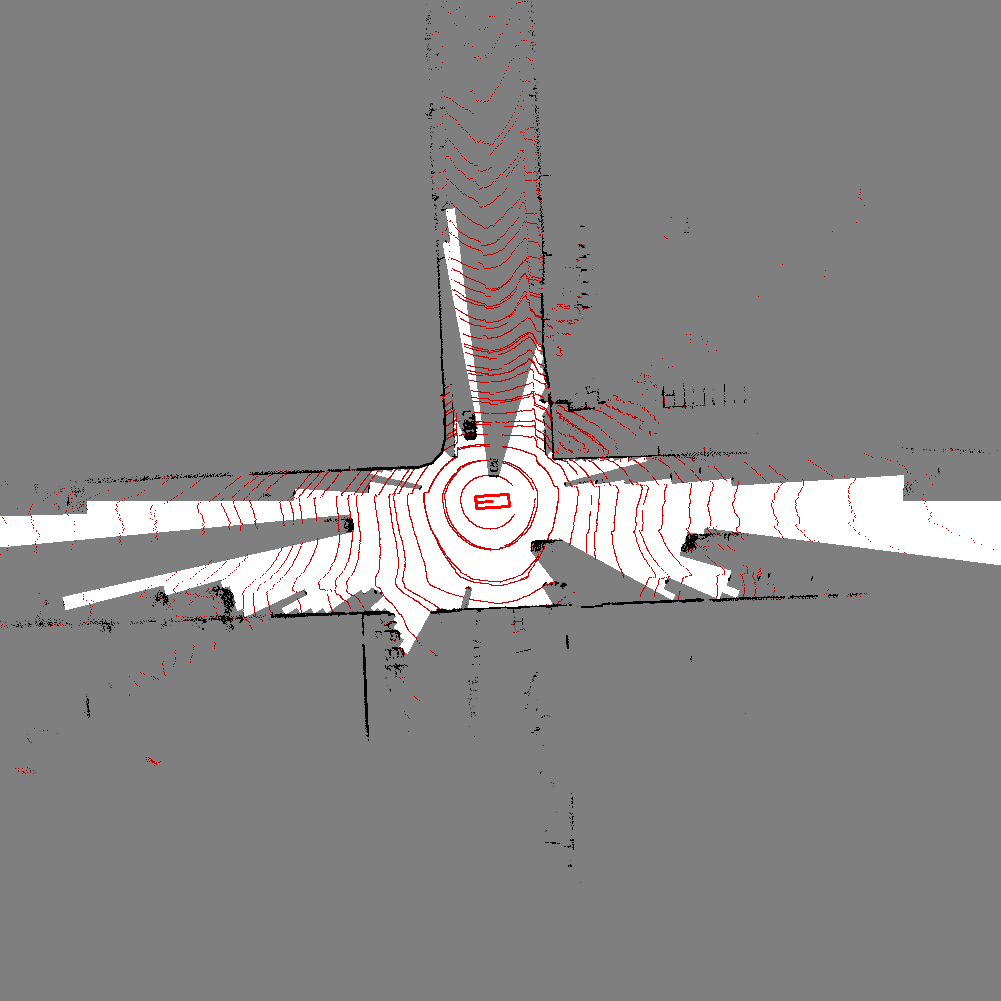}
	\end{subfigure}
	\begin{subfigure}{0.32\columnwidth}
	\includegraphics[trim=10cm 13cm 10cm 6cm, clip, width=\columnwidth]{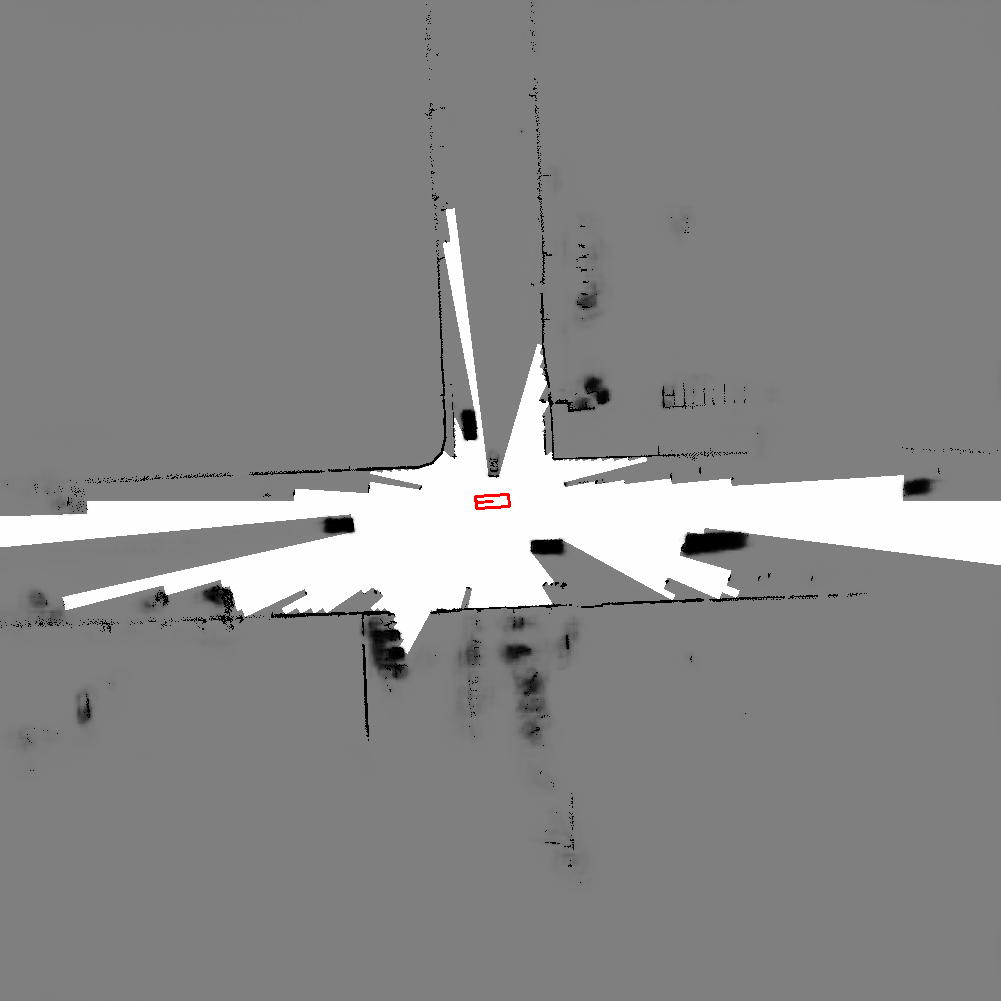}
	\end{subfigure}
	\begin{subfigure}{0.32\columnwidth}
	\includegraphics[trim=10cm 13cm 10cm 6cm, clip,width= \columnwidth]{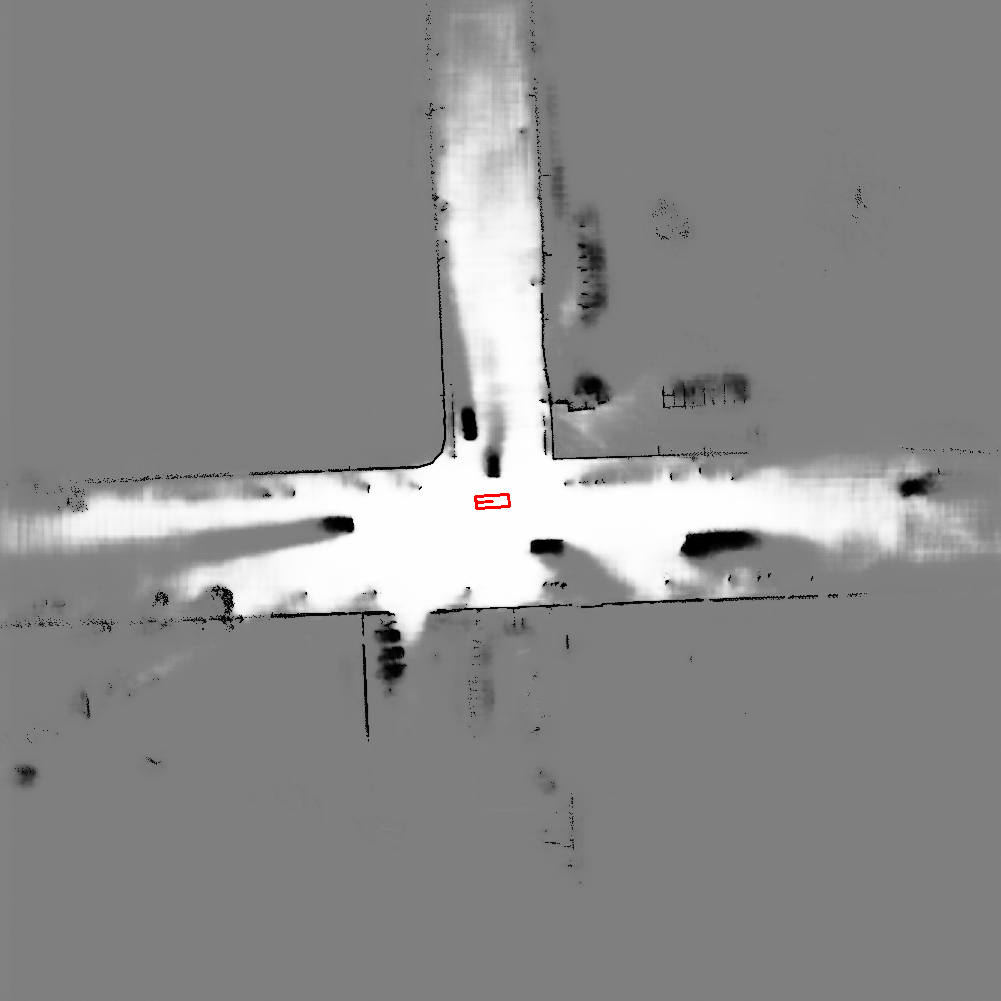}
	\end{subfigure}
	\caption{In the left image a measurement grid map based on a geometric inverse sensor model is visualized, with lidar ground measurements shown in red. The occupancy grid map prediction of the configuration \textit{meas\_base} is depicted in the middle, the prediction of \textit{lidar\_base\_gc} in the right image.}
	\label{fig:EvaluateOccupancy}
	\vspace{-0.5cm}
\end{figure}
In this section, we compare the representation of occupancy and freespace in a classical occupancy grid map, i.e. the measurement grid maps described in Section \ref{sec:MeasGrids} with the predictions of our network architecture.
In contrast to classical occupancy grid maps, our approach has the ability to predict the shape of objects as occupied area, as visualized in Fig.~\ref{fig:EvaluateOccupancy}.
In the measurement grid map, the vehicles are represented with the occupied cells, that are obtained by the sparse lidar measurements. 
However, both network configurations, using measurement grid maps or raw lidar data as input, predict a rectangle area as occupied, which is an improved modeling of the real world occupied space.
In the depicted scenario an object near to the ego-vehicle causes a large occlusion in the measurement grid map, even though there are ground measurements visible in this area.
The network with the setting \textit{meas\_base} simply predicts the freespace as it is provided at the network input and therefore has the same limitations as the geometric inverse sensor model.
In contrast, the network using the raw lidar data learns an improved inverse sensor model and make use of the ground points to predict freespace.
\subsection{Quantitative Evaluation}
\begin{table*}[]
	\vspace*{1.42mm}
	\renewcommand{\arraystretch}{1.2}
	\centering
	\caption{Quantitative evaluation of the velocity prediction and semantic classification. The semantic classes are VEHICLE (V), LARGE VEHICLE (LV), PEDESTRIAN (PED), TWO WHEELER (TW) and STATIC (STA).}
	\label{tab:QuantitativeEval}
	\setlength{\tabcolsep}{0.2cm}
	\begin{tabular}{|c||ccc||cccccccccc|}
		\hline
		task  & \multicolumn{3}{c||}{velocity}                                                           & \multicolumn{10}{c|}{semantic}                                                                                                  \\ \hline
		\diagbox{method}{metric} & \multicolumn{1}{c|}{mIoU} & EPE\ts{occ} & \multicolumn{1}{c||}{EPE\ts{dyn}} & V   &  LV  & PED & TW  & STA & V\ts{static} & LV\ts{static} & PED\ts{static} & \multicolumn{1}{c|}{TW\ts{static}} & mIoU \\ \hline
		meas\_base           & \multicolumn{1}{c|}{92.85}& \tbf{0.0213}  & \tbf{0.1147} & 84.51& \tbf{60.40} & 62.08 & 28.87 & 87.96 &  86.27  & 57.25  & 5.10 & \multicolumn{1}{c|}{0.14}        & 52.51\\
		lidar\_base          & \multicolumn{1}{c|}{93.08}& 0.0220  & 0.1285      & 82.98 & 52.56 & 66.69 & 20.78 & 88.80 & 89.13 & 52.86  &  14.67 & \multicolumn{1}{c|}{\tbf{3.00}}  & 52.39 \\
		lidar\_base\_gc      & \multicolumn{1}{c|}{93.07} & 0.0229   & 0.1333    & \tbf{85.70} & 57.68 & \tbf{69.45} & 26.59  & \tbf{89.95} & \tbf{90.49} & \tbf{64.58}  & \tbf{15.82}  & \multicolumn{1}{c|}{0.83}   & \tbf{55.68}\\
		lidar\_base\_road      & \multicolumn{1}{c|}{92.69} & 0.0244   &  0.1434 & 83.96 & 55.60 & 62.27 & \tbf{35.54}  & 89.19  & 90.18 & 50.98  & 10.73  & \multicolumn{1}{c|}{0.65}  & 53.23 \\	
		lidar\_base\_road\_gc  & \multicolumn{1}{c|}{\tbf{93.40}} & 0.0221   &  0.1300 & 84.05 & 60.31 & 65.58 & 32.33 & 89.05 & 89.55 & 58.10 		&  15.79 & \multicolumn{1}{c|}{2.82}   &  55.29  \\ \hline
	\end{tabular}
	\vspace*{-0.2cm}
\end{table*}
For the evaluation of the velocity predictions, we calculate the end-point-error (EPE), which is the L2 distance between the estimated 2D velocity vector and the ground truth.
We report two metrics, considering all occupied cells (EPE\ts{occ}) and all dynamic cells (EPE\ts{dyn}), i.e. cells with a ground truth velocity magnitude $v\ts{mag} > 0.8~\text{m/s}$.
The same threshold is used to separate the occupied cells in the two classes static or dynamic.
The performance of the separation in static and dynamic cells is then measured by calculating the mean intersection over union (mIoU).
For the evaluation of the semantic classification, we provide the intersection over union (IoU) for each class and the mIoU as overall performance measurement. 
The results in Table~\ref{tab:QuantitativeEval} show, that the configuration, using the measurement grid maps as input data \textit{meas\_base} provides marginally more accurate velocity predictions compared to \textit{lidar\_base}.
We argue, that with the \textit{meas\_base} configuration the label data is similar to the input data, so there is more network capacity for the other tasks available, which leads to slightly better results.
The best results for the semantic classification are obtained using the lidar input data and the auxiliary task $y\ts{gc}$.
It is notable, that the \textit{meas\_base} configuration achieves comparable results for the semantic classification of the dynamic classes, as these objects are easier separable from the static environment due to the recurrent network architecture.
However, especially for static and small objects, e.g. pedestrians, the configurations with the raw lidar input data provide improved results.
\subsection{Runtime Performance}
\begin{table}
	\renewcommand{\arraystretch}{1.2}
	\centering
	\caption{Runtime performance in ms.}
	\label{tab:inferenceTimes}
	\setlength{\tabcolsep}{0.1cm}
	\begin{tabular}{|cc||cc||c|}
		\hline
		method & sensor setup & preprocessing & network graph & overall \\
		\hline
		meas\_base		& VLP32 & 25 & 8 & 33 \\
		lidar\_base     & VLP32 & 5 & 13 & 18 \\
		lidar\_base\_road & VLP32 & 5 & 14 & 19 \\
		lidar\_base\_road & HDL64 & 9 & 14 & 23 \\
		\hline
	\end{tabular}
	\vspace{-0.5cm}
\end{table}
We integrated our multi-task network in the ROS framework for deployment with data from our experimental vehicle.
In contrast to the usage of input sequences in the training setting, the network graph for deployment only use the current input data and the recurrent states of the previous time step to predict the current grid map.
So, the network graph for the deployment is saved for inputs with sequence length $n\ts{in} = 1$ and for the complete grid map with $1001\times1001$ cells, leading to a perceived area of $150\text{m}\times150\text{m}$. 
The preprocessing of the lidar data is executed on the CPU, whereas the network graph runs on an Nvidia GeForce RTX 2080Ti using half-precision floating mode. 
We compare the runtime performance of our network relying on measurement grid maps as input \textit{meas\_base} with our end-to-end approach \textit{lidar\_base} in Table \ref{tab:inferenceTimes}, using data from a 32 layer lidar.
The preprocessing of the \textit{meas\_base} configuration consists of a ground point removal step and the calculation of measurement grid maps, whereas the end-to-end approach only needs the BEV projection described in Section \ref{sec:InputData}.
Therefore, the configuration \textit{lidar\_base} achieves a lower overall runtime, however both configurations are real-time capable.
Note, that the specified time for the network graph includes the time for transferring the data to and from the GPU.
Additionally, we apply the configuration \textit{lidar\_base\_road} with the data of a 64 layer lidar, which results in an increased preprocessing time, but the inference time of the network graph remains constant.
This result shows, that our approach is also usable in real-time with lidar sensors that provide larger point clouds.
\subsection{Qualitative Evaluation}
\begin{figure}
	\centering	
	\begin{subfigure}{\columnwidth}
		\includegraphics[trim=5cm 0cm 7cm 5cm, clip,width=0.49\columnwidth]{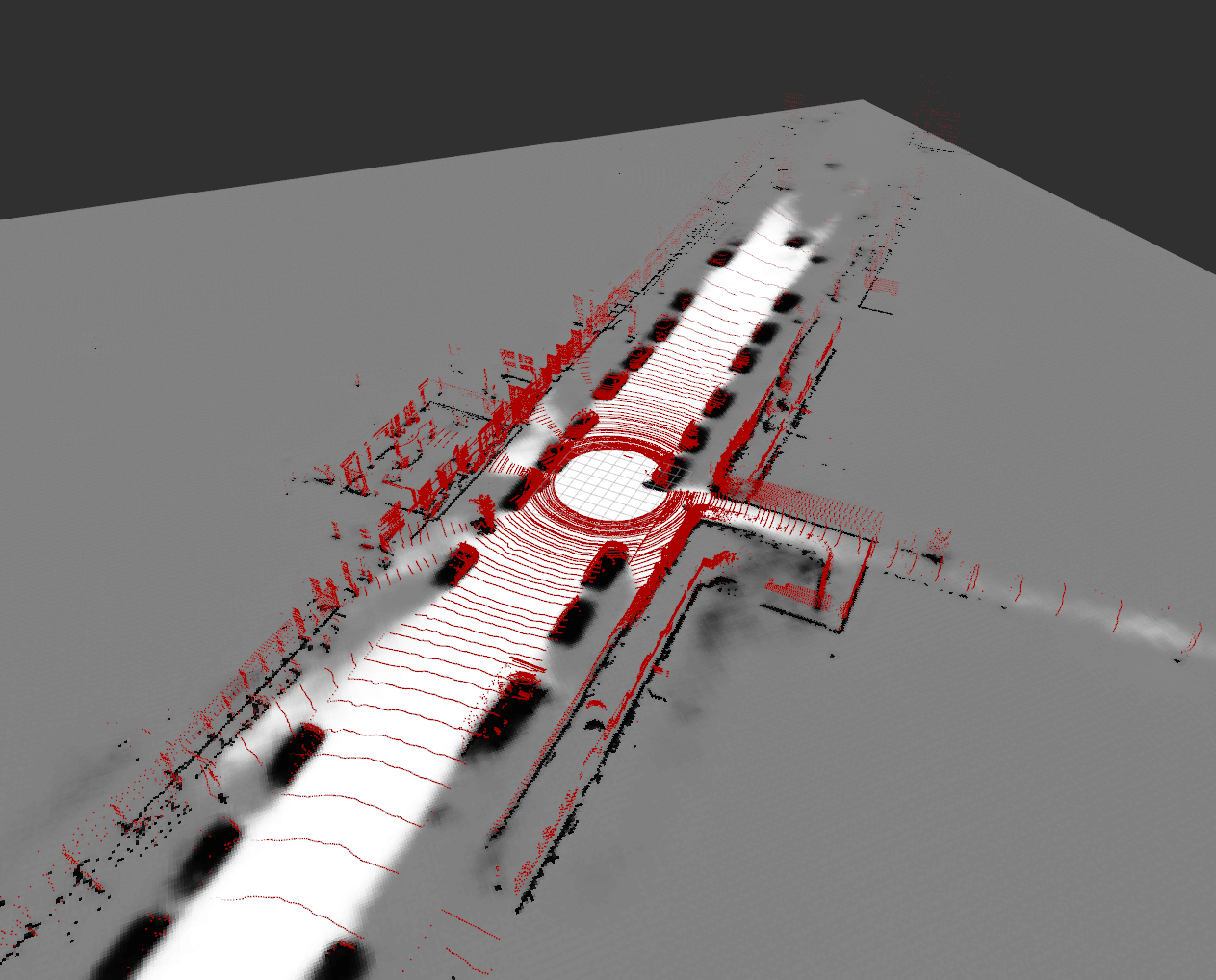}
		\includegraphics[trim=5cm 0cm 7cm 5cm, clip,width=0.49\columnwidth]{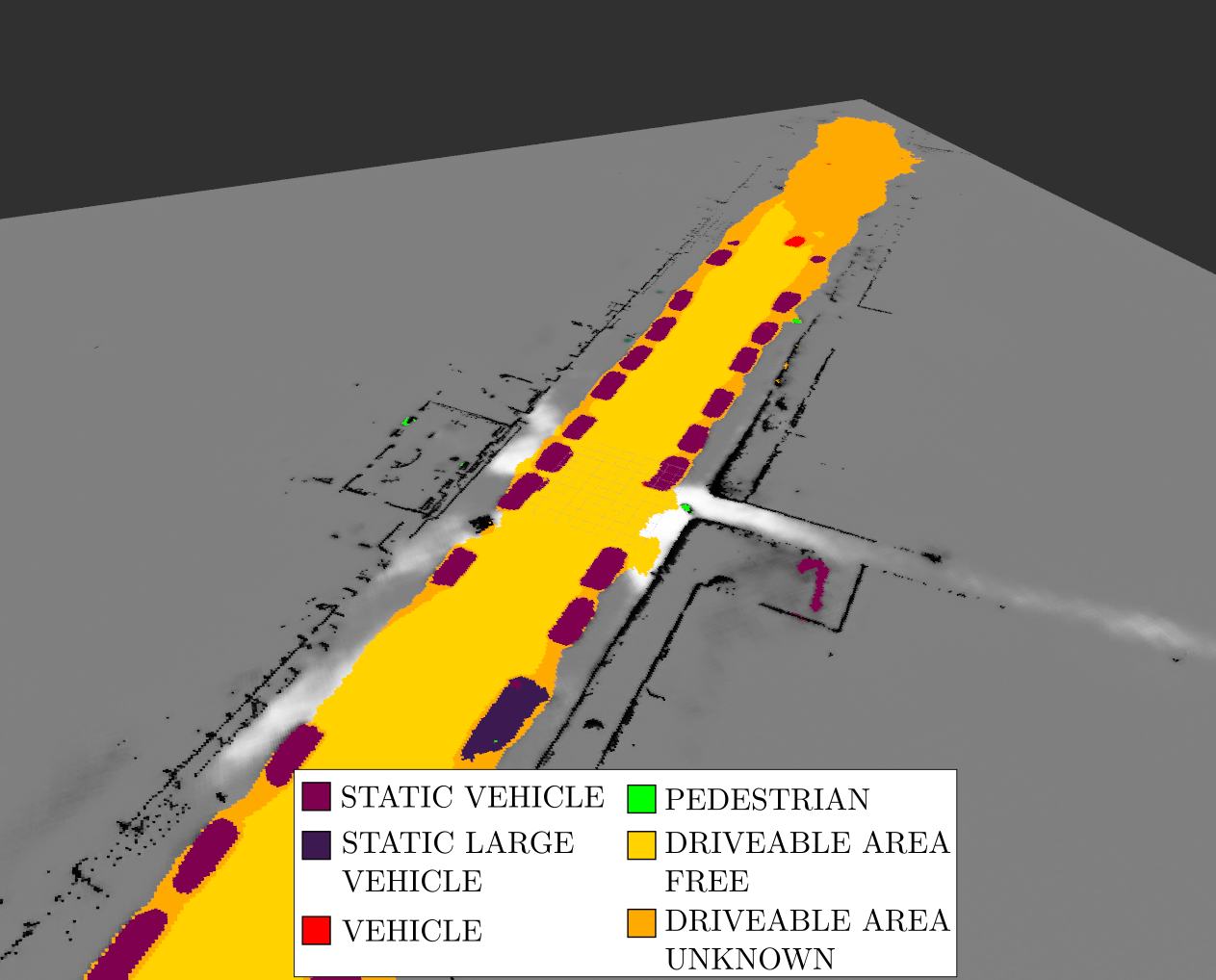}
		\caption{Left: Occupancy grid map with lidar points. Right: Semantic grid map with driveable area.}
		\label{fig:DeployROSa}
		\vspace{0.1cm}
	\end{subfigure}
	\begin{subfigure}{\columnwidth}
		\includegraphics[trim=8cm 0cm 4cm 5cm, clip,width=0.49\columnwidth]{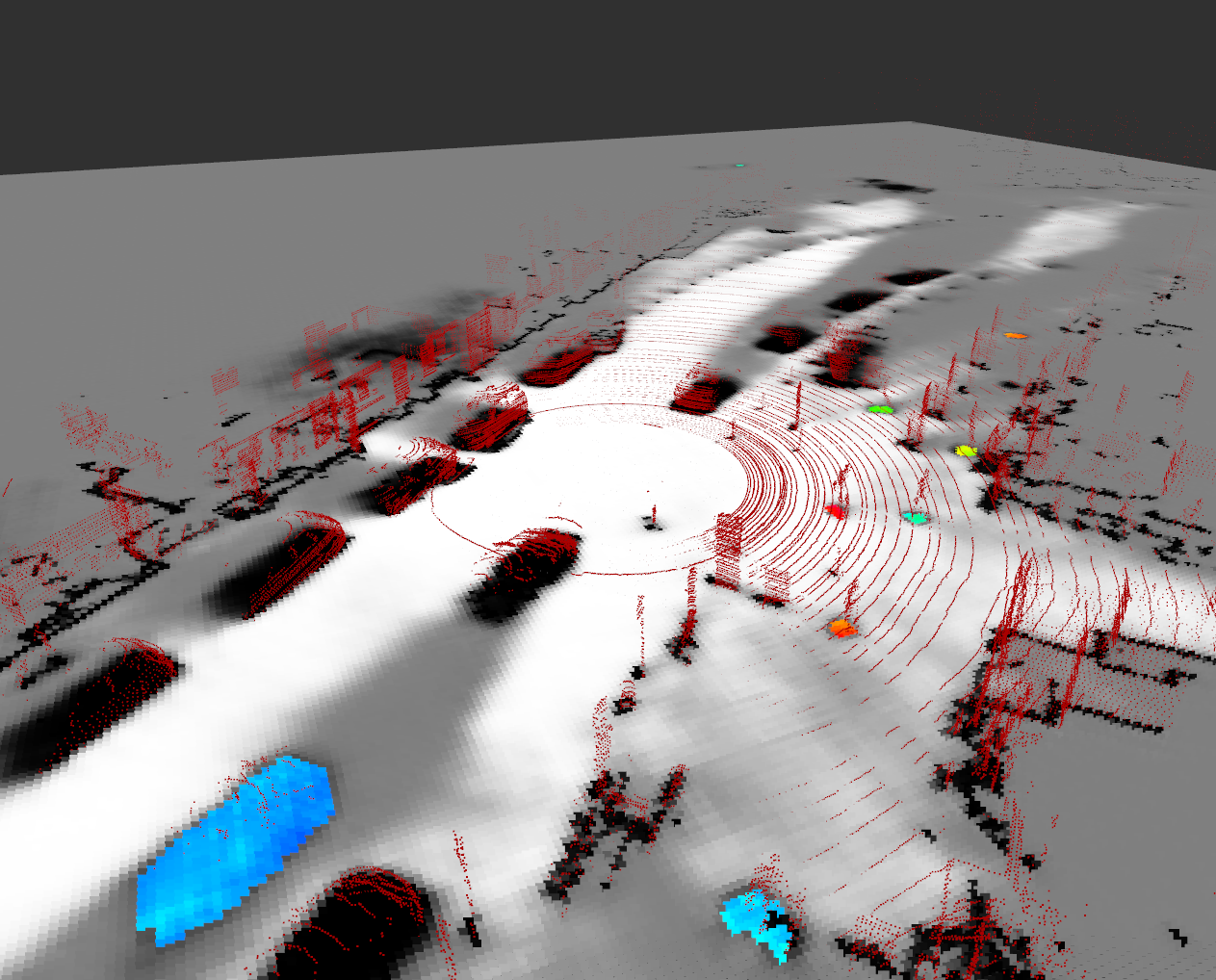}
		\includegraphics[trim=8cm 0cm 4cm 5cm, clip,width=0.49\columnwidth]{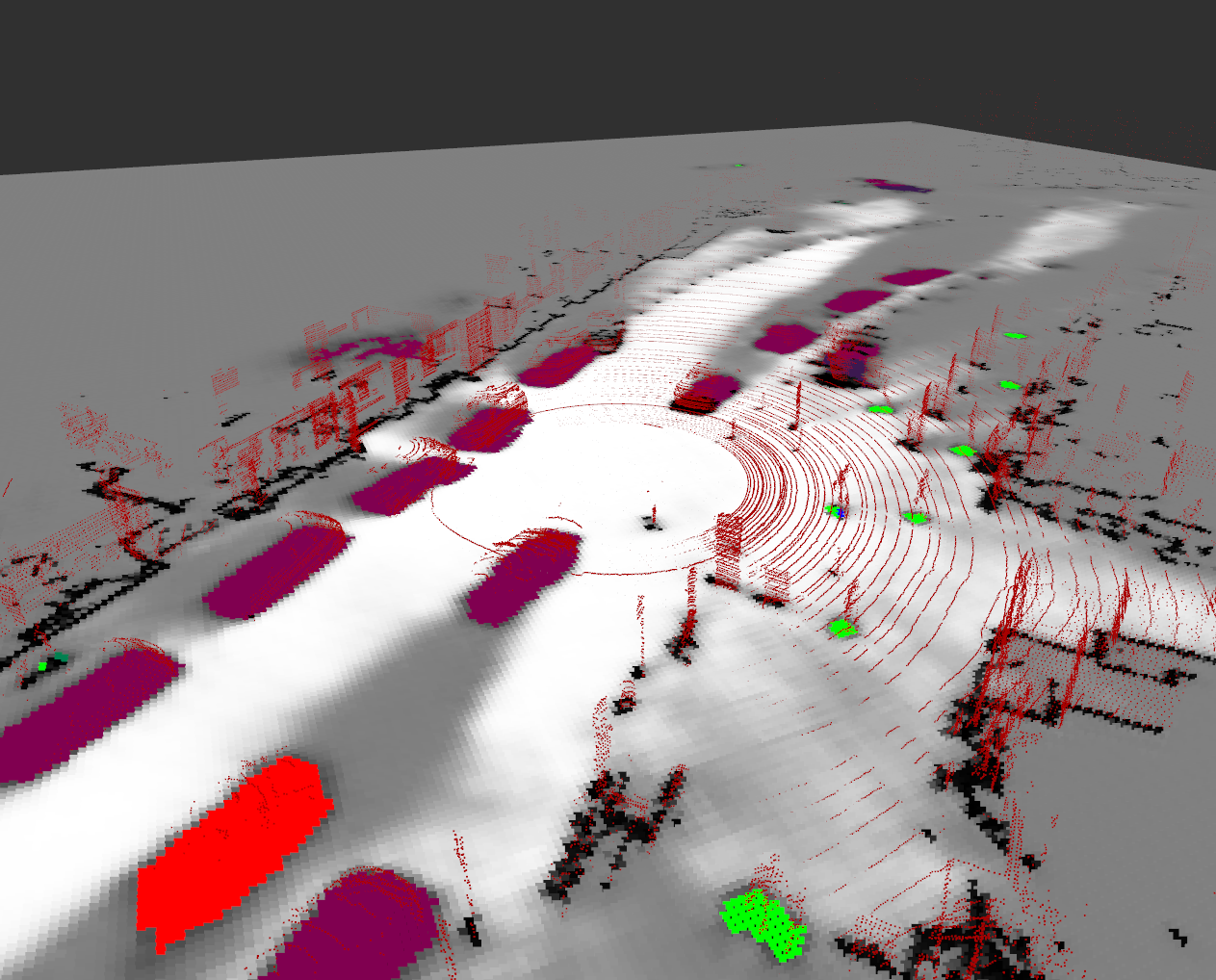}
		\caption{Left: Dynamic occupancy grid map with dynamic cells visualized with colors according to the velocity orientation. Right: Semantic grid map with colors for the semantic classes.}
		\label{fig:DeployROSb}
	\end{subfigure}
	\caption{Illustration of a scenario with parking vehicles and straight lane (a) and a scenario with several pedestrians (b). Best viewed digitally with zoom.}
	\label{fig:DeployROS}
	\vspace{-0.5cm}
\end{figure}
In this section, we show some qualitative results of our approach with the configuration \textit{lidar\_base\_road\_gc}, using data obtained by our experimental vehicle, equipped with a 64 layer lidar scanner.
In Fig.~\ref{fig:DeployROSa} the ego-vehicle drives on a straight roadway with several parking vehicles.
The visualization on the left depicts the predicted occupancy grid map with the lidar points in red and shows the ability of the network to predict freespace in areas with ground measurements.
In the right image the semantic classification and the prediction of the driveable area are shown in colors.
The network correctly predicts the parking vehicles (purple) and the large static vehicle (dark purple), as well as the dynamic vehicle (red) and the pedestrian (green) at the top right of the image.
In our experiments, we observed that the driveable area prediction generalizes not so well to other sensor setups or new scenarios. 
So it mainly works well in scenes with straight roads, parking vehicles or visible curbs. 
However, this problem could be solved using the same sensor data in the training setup and a more diverse dataset.
In the scenario depicted in Fig.~\ref{fig:DeployROSb}, the ego-vehicle has stopped behind a line of cars, with several pedestrians walking on the right side.
The left image shows the dynamic occupancy grid map, with colors indicating the direction of the velocity estimates for dynamic cells.
The network correctly predicts the moving cells as pedestrians, shown in green in the right image. 
In such scenarios with small objects, our approach benefits from the recurrent network architecture as it can use information of multiple time steps to detect pedestrians based on their motion.
\section{Conclusions}
%
%
In this work, we have presented a multi-task recurrent neural network to predict dynamic occupancy grid maps with semantic classes.
Our model uses raw 3D lidar data, represented as BEV image with several height channels as input data and outputs an occupancy grid map. 
Compared to prior work, this approach does not need a ground point removal and measurement grid map calculation as preprocessing steps and therefore achieves a lower runtime performance.
Moreover, our experiments show an improved performance for the semantic classification, especially for static objects.
In future work, we plan to investigate the possibility of adding additional tasks, such as the prediction of instances.
%
%
%
%



%
%
%
%
%
%
\bibliographystyle{IEEEtran}
\bibliography{IEEEtranControl,IEEEabrv,mybibfile}

\begin{thebibliography}{10}
\providecommand{\url}[1]{#1}
\csname url@rmstyle\endcsname
\providecommand{\newblock}{\relax}
\providecommand{\bibinfo}[2]{#2}
\providecommand\BIBentrySTDinterwordspacing{\spaceskip=0pt\relax}
\providecommand\BIBentryALTinterwordstretchfactor{4}
\providecommand\BIBentryALTinterwordspacing{\spaceskip=\fontdimen2\font plus
\BIBentryALTinterwordstretchfactor\fontdimen3\font minus
  \fontdimen4\font\relax}
\providecommand\BIBforeignlanguage[2]{{%
\expandafter\ifx\csname l@#1\endcsname\relax
\typeout{** WARNING: IEEEtran.bst: No hyphenation pattern has been}%
\typeout{** loaded for the language `#1'. Using the pattern for}%
\typeout{** the default language instead.}%
\else
\language=\csname l@#1\endcsname
\fi
#2}}

\bibitem{TanzmeisterGridBasedMapping}
G.~{Tanzmeister}, J.~{Thomas}, D.~{Wollherr}, and M.~{Buss}, ``Grid-based
  mapping and tracking in dynamic environments using a uniform evidential
  environment representation,'' in \emph{2014 IEEE International Conference on
  Robotics and Automation (ICRA)}, May 2014, pp. 6090--6095.

\bibitem{Negre_HybridSamplingBayesian}
A.~{N\`{e}gre}, L.~{Rummelhard}, and C.~{Laugier}, ``Hybrid sampling bayesian
  occupancy filter,'' in \emph{2014 IEEE Intelligent Vehicles Symposium
  Proceedings}, June 2014, pp. 1307--1312.

\bibitem{DBLP:journals/corr/NussRTYKMGD16}
D.~Nuss, S.~Reuter, M.~Thom, T.~Yuan, G.~Krehl, M.~Maile, A.~Gern, and
  K.~Dietmayer, ``A random finite set approach for dynamic occupancy grid maps
  with real-time application,'' \emph{The International Journal of Robotics
  Research}, vol.~37, no.~8, pp. 841--866, 2018.

\bibitem{Thrun:2005:PR:1121596}
S.~Thrun, W.~Burgard, and D.~Fox, \emph{Probabilistic Robotics (Intelligent
  Robotics and Autonomous Agents)}.\hskip 1em plus 0.5em minus 0.4em\relax
  Cambridge, Mass.: MIT Press, 2005.

\bibitem{SchreiberDynamicOccupancyGridMapping}
M.~Schreiber, V.~Belagiannis, C.~Gl\"aser, and K.~Dietmayer, ``Dynamic
  occupancy grid mapping with recurrent neural networks,'' in \emph{2021 IEEE
  International Conference on Robotics and Automation (ICRA)}, 2021, pp.
  6717--6724.

\bibitem{ElfesOccGrids}
A.~{Elfes}, ``Using occupancy grids for mobile robot perception and
  navigation,'' \emph{Computer}, vol.~22, no.~6, pp. 46--57, June 1989.

\bibitem{DanescuParticleBasedOccupancyGrid}
R.~{Danescu}, F.~{Oniga}, and S.~{Nedevschi}, ``Modeling and tracking the
  driving environment with a particle-based occupancy grid,'' \emph{IEEE
  Transactions on Intelligent Transportation Systems}, vol.~12, no.~4, pp.
  1331--1342, Dec 2011.

\bibitem{SchreiberMotionEstimationInOccupancyGrids}
M.~{Schreiber}, V.~{Belagiannis}, C.~{Gl\"aser}, and K.~{Dietmayer}, ``Motion
  estimation in occupancy grid maps in stationary settings using recurrent
  neural networks,'' in \emph{2020 IEEE International Conference on Robotics
  and Automation (ICRA)}, 2020, pp. 8587--8593.

\bibitem{AnyMotionDetector}
A.~{Filatov}, A.~{Rykov}, and V.~{Murashkin}, ``Any motion detector: Learning
  class-agnostic scene dynamics from a sequence of lidar point clouds,'' in
  \emph{2020 IEEE International Conference on Robotics and Automation (ICRA)},
  2020, pp. 9498--9504.

\bibitem{Zhou2018VoxelNetEL}
Y.~Zhou and O.~Tuzel, ``Voxelnet: End-to-end learning for point cloud based 3d
  object detection,'' \emph{2018 IEEE/CVF Conference on Computer Vision and
  Pattern Recognition}, pp. 4490--4499, 2018.

\bibitem{lee2020pillarflow}
K.-H. Lee, M.~Kliemann, A.~Gaidon, J.~Li, C.~Fang, S.~Pillai, and W.~Burgard,
  ``Pillarflow: End-to-end birds-eye-view flow estimation for autonomous
  driving,'' 2020.

\bibitem{lang2019pointpillars}
A.~H. Lang, S.~Vora, H.~Caesar, L.~Zhou, J.~Yang, and O.~Beijbom,
  ``Pointpillars: Fast encoders for object detection from point clouds,'' in
  \emph{Proceedings of the IEEE Conference on Computer Vision and Pattern
  Recognition}, 2019, pp. 12\,697--12\,705.

\bibitem{MotionNet}
P.~{Wu}, S.~{Chen}, and D.~N. {Metaxas}, ``Motionnet: Joint perception and
  motion prediction for autonomous driving based on bird’s eye view maps,''
  in \emph{2020 IEEE/CVF Conference on Computer Vision and Pattern Recognition
  (CVPR)}, 2020, pp. 11\,382--11\,392.

\bibitem{ThrunExplorationAndModelBuilding}
S.~Thrun, ``Exploration and model building in mobile robot domains,'' in
  \emph{IEEE International Conference on Neural Networks}, 1993, pp. 175--180
  vol.1.

\bibitem{DequaireDeepTrackingInTheWild}
J.~Dequaire, P.~Ondr\'{u}\v{s}ka, D.~Rao, D.~Wang, and I.~Posner, ``Deep
  tracking in the wild: End-to-end tracking using recurrent neural networks,''
  \emph{The International Journal of Robotics Research}, vol.~37, no. 4-5, pp.
  492--512, 2018.

\bibitem{WirgesEvidentialOccupancyGridMapAugmentation}
S.~Wirges, C.~Stiller, and F.~Hartenbach, ``Evidential occupancy grid map
  augmentation using deep learning,'' in \emph{2018 IEEE Intelligent Vehicles
  Symposium (IV)}, 2018, pp. 668--673.

\bibitem{VanKempenASimulationBasedEndToEndLearning}
R.~Van~Kempen, B.~Lampe, T.~Woopen, and L.~Eckstein, ``A simulation-based
  end-to-end learning framework for evidential occupancy grid mapping,'' in
  \emph{2021 IEEE Intelligent Vehicles Symposium (IV)}, 2021, pp. 934--939.

\bibitem{LuMonocularSemanticOccupancyGridMapping}
C.~Lu, M.~J.~G. van~de Molengraft, and G.~Dubbelman, ``Monocular semantic
  occupancy grid mapping with convolutional variational encoder–decoder
  networks,'' \emph{IEEE Robotics and Automation Letters}, vol.~4, no.~2, pp.
  445--452, 2019.

\bibitem{RoddickPredictingSemanticMapRepresentations}
T.~Roddick and R.~Cipolla, ``Predicting semantic map representations from
  images using pyramid occupancy networks,'' in \emph{2020 IEEE/CVF Conference
  on Computer Vision and Pattern Recognition (CVPR)}, 2020, pp.
  11\,135--11\,144.

\bibitem{ErkentSemanticGridEstimationWithAHybrid}
O.~Erkent, C.~Wolf, C.~Laugier, D.~S. Gonzalez, and V.~R. Cano, ``Semantic grid
  estimation with a hybrid bayesian and deep neural network approach,'' in
  \emph{2018 IEEE/RSJ International Conference on Intelligent Robots and
  Systems (IROS)}, 2018, pp. 888--895.

\bibitem{RummelhardCMCDOT}
L.~Rummelhard, A.~Nègre, and C.~Laugier, ``Conditional monte carlo dense
  occupancy tracker,'' in \emph{2015 IEEE 18th International Conference on
  Intelligent Transportation Systems}, 2015, pp. 2485--2490.

\bibitem{BiederExploitingMultiLayerGridMapsForSurroundViewSemantic}
F.~Bieder, S.~Wirges, J.~Janosovits, S.~Richter, Z.~Wang, and C.~Stiller,
  ``Exploiting multi-layer grid maps for surround-view semantic segmentation of
  sparse lidar data,'' in \emph{2020 IEEE Intelligent Vehicles Symposium (IV)},
  2020, pp. 1892--1898.

\bibitem{DeepLab}
L.-C. Chen, G.~Papandreou, I.~Kokkinos, K.~Murphy, and A.~L. Yuille, ``Deeplab:
  Semantic image segmentation with deep convolutional nets, atrous convolution,
  and fully connected crfs,'' \emph{IEEE Transactions on Pattern Analysis and
  Machine Intelligence}, vol.~40, no.~4, pp. 834--848, 2018.

\bibitem{BiederImprovingLidarBasedSemanticSegmentationOfTopViewGridMaps}
F.~Bieder, M.~Link, S.~Romanski, H.~Hu, and C.~Stiller, ``Improving lidar-based
  semantic segmentation of top-view grid maps by learning features in
  complementary representations,'' in \emph{2021 IEEE 24th International
  Conference on Information Fusion (FUSION)}, 2021, pp. 1--7.

\bibitem{MilitoRangeNet++}
A.~Milioto, I.~Vizzo, J.~Behley, and C.~Stachniss, ``Rangenet ++: Fast and
  accurate lidar semantic segmentation,'' in \emph{2019 IEEE/RSJ International
  Conference on Intelligent Robots and Systems (IROS)}, 2019, pp. 4213--4220.

\bibitem{FeiPillarSegNet}
J.~Fei, K.~Peng, P.~Heidenreich, F.~Bieder, and C.~Stiller, ``Pillarsegnet:
  Pillar-based semantic grid map estimation using sparse lidar data,'' in
  \emph{2021 IEEE Intelligent Vehicles Symposium (IV)}, 2021, pp. 838--844.

\bibitem{YangPIXOR}
B.~Yang, W.~Luo, and R.~Urtasun, ``Pixor: Real-time 3d object detection from
  point clouds,'' in \emph{2018 IEEE/CVF Conference on Computer Vision and
  Pattern Recognition}, 2018, pp. 7652--7660.

\bibitem{Argoverse}
M.-F. Chang, J.~W. Lambert, P.~Sangkloy, J.~Singh, S.~Bak, A.~Hartnett,
  D.~Wang, P.~Carr, S.~Lucey, D.~Ramanan, and J.~Hays, ``Argoverse: 3d tracking
  and forecasting with rich maps,'' in \emph{Conference on Computer Vision and
  Pattern Recognition (CVPR)}, 2019.

\bibitem{DBLP:journals/corr/RonnebergerFB15}
O.~Ronneberger, P.~Fischer, and T.~Brox, ``U-net: Convolutional networks for
  biomedical image segmentation,'' in \emph{Medical Image Computing and
  Computer-Assisted Intervention}, 2015, pp. 234--241.

\bibitem{DBLP:journals/corr/ShiCWYWW15}
X.~SHI, Z.~Chen, H.~Wang, D.-Y. Yeung, W.-k. Wong, and W.-c. WOO,
  ``Convolutional lstm network: A machine learning approach for precipitation
  nowcasting,'' in \emph{Advances in Neural Information Processing Systems
  (NIPS) 28}.\hskip 1em plus 0.5em minus 0.4em\relax Curran Associates, Inc.,
  2015, pp. 802--810.

\bibitem{LinFocalLoss}
T.-Y. Lin, P.~Goyal, R.~Girshick, K.~He, and P.~Dollár, ``Focal loss for dense
  object detection,'' in \emph{2017 IEEE International Conference on Computer
  Vision (ICCV)}, 2017, pp. 2999--3007.

\bibitem{DBLP:journals/corr/KingmaB14}
\BIBentryALTinterwordspacing
D.~P. {Kingma} and J.~{Ba}, ``{Adam: A Method for Stochastic Optimization},''
  \emph{arXiv preprint arXiv:1412.6980}, 2014. [Online]. Available:
  \url{http://arxiv.org/abs/1412.6980}
\BIBentrySTDinterwordspacing

\bibitem{DBLP:journals/corr/ZarembaSV14}
\BIBentryALTinterwordspacing
W.~{Zaremba}, I.~{Sutskever}, and O.~{Vinyals}, ``{Recurrent Neural Network
  Regularization},'' \emph{arXiv preprint arXiv:1409.2329}, 2014. [Online].
  Available: \url{http://arxiv.org/abs/1409.2329}
\BIBentrySTDinterwordspacing

\end{thebibliography}
\end{document}